# Primal View on Belief Propagation

Tomáš Werner

Center for Machine Perception, Czech Technical University
Karlovo náměstí 13, 12135 Prague, Czech Republic

## Abstract

It is known that fixed points of loopy belief propagation (BP) correspond to stationary points of the Bethe variational problem, where we minimize the Bethe free energy subject to normalization and marginalization constraints. Unfortunately, this does not entirely explain BP because BP is a dual rather than primal algorithm to solve the Bethe variational problem – beliefs are infeasible before convergence. Thus, we have no better understanding of BP than as an algorithm to seek for a common zero of a system of non-linear functions, not explicitly related to each other. In this theoretical paper, we show that these functions are in fact explicitly related – they are the partial derivatives of a single function of reparameterizations. That means, BP seeks for a stationary point of a single function, without any constraints. This function has a very natural form: it is a linear combination of local log-partition functions, exactly as the Bethe entropy is the same linear combination of local entropies.

## 1 Introduction

Loopy belief propagation (further only belief propagation, BP) (Pearl, 1988) is a well-known algorithm to approximate marginals and the partition function of the Gibbs probability distribution defined by an undirected graphical model (Markov random field). For acyclic graphs it yields the exact result, for graphs with cycles it often yields surprisingly good approximations. A large body of literature exists on BP and related topics and we refer the reader to the recent survey by (Wainwright & Jordan, 2008).

Unfortunately, BP on cyclic graphs is not guaranteed to converge, which is indeed often observed. A lot of effort has been invested to understanding this phenomenon, see (Wainwright & Jordan, 2008, §4.1.3) for references. Solid ground was provided by (Yedidia et al., 2000; Yedidia et al., 2005) who discovered that BP fixed points coincide with stationary points of the Bethe variational problem, long known in statistical physics. (Heskes, 2003; Heskes, 2006) showed that every stable BP fixed points are local optima (rather than saddle points) of this problem, but not *vice versa*.

The basic operation in the BP algorithm is 'passing a message', which means sending a vector of numbers between a node and an edge of the graph (Pearl, 1988). Messages turned out to be directly related to the Lagrange multipliers of the Bethe variational problem (Yedidia et al., 2000; Yedidia et al., 2005). Later it became clear (Wainwright et al., 2004) that passing a message corresponds to reparameterizing the distribution. In this view, BP tries to reparameterize the distribution so that the corresponding beliefs have consistent marginals.

Though this is generally known, no existing theoretical analysis of BP fully utilizes the interpretation of messages and the Lagrange multipliers of the Bethe variational problem as reparameterizations. In contrast, we incorporate reparameterizations into variational inference and BP in a principled way, which makes the picture more complete and provides a mathematical framework in which we formulate our main result.

The correspondence between BP fixed points and the Bethe variational problem did not entirely explain the BP algorithm itself because BP does not directly solve the Bethe variational problem – beliefs are infeasible to this problem until convergence. Thus, we still have no better understanding of BP than an algorithm to seek for a common zero of a system of non-linear functions, not explicitly related to each other. As our main result, we show that these functions are in fact explicitly related – they are the partial derivatives of a single function of reparameterizations. In other words, BP searches for a stationary point of a single func-tion, without any constraints. This function has a very natural form: it is a linear combination of local log-partition functions, exactly as the Bethe entropy is the same linear combination of local entropies. We show that BP fixed points are in one-to-one correspondence with stationary points of this function and that all these points are saddles[1].

Several versions of BP and related free energies exist. Originally, BP was formulated for models with pairwise interactions. We formulate our result for the factor-graph BP (Kschischang et al., 2001), which permits interactions of arbitrary order. We currently do not consider more complex versions, cluster variation methods and generalized BP (Yedidia et al., 2005).

## 2 Exponential families

Here we recall the basics of exponential families of probability distributions, which offer a convenient formalism to reason about graphical models (Wainwright & Jordan, 2008). In §2.3, we incorporate the concept of *reparameterizations* in overcomplete exponential families in a more principled way than other authors – which is important for graphical models, where reparameterizations play a crucial rôle.

Let $X$ and $I$ be finite sets and $\phi \colon X \to \mathbb{R}^I$. The discrete exponential family is a family of probability distributions

$$p(x|\theta) = \exp[\theta\phi(x) - F(\theta)] \qquad (1)$$

instantiated by triplet $(X, I, \phi)$ and parameterized by vector $\theta \in \mathbb{R}^I$. We understand $\theta$ as a row vector and $\phi(x)$ as a column vector, so that $\theta\phi(x) = \sum_{i \in I} \theta_i \phi_i(x)$. The normalization term

$$F(\theta) = \bigoplus_{x \in X} \theta\phi(x) \qquad (2)$$

is the *log-partition function* and $a \oplus b = \log(e^a + e^b)$ denotes the *log-sum-exp operation*. Operation $\oplus$ is associative and commutative and $+$ distributes over $\oplus$.

We assume in §2.1 and §2.2 that the functions $\phi_i$ are affinely independent, i.e., they form a *minimal representation* of the family. We relax this later in §2.3.

### 2.1 Mean parameters

The mean values of functions $\phi_i$ with respect to distribution (1) are the *mean parameters* of the distribution

$$\mu = \sum_{x \in X} p(x|\theta)\phi(x) = \frac{\sum_{x \in X} \phi(x)\exp\theta\phi(x)}{\sum_{x \in X} \exp\theta\phi(x)} \qquad (3)$$

---

[1] These saddle points should not be confused with the saddle points in the double-loop algorithms to minimize the Bethe free energy (Heskes, 2003; Heskes, 2006).

which is a column vector. The map $\theta \mapsto \mu$ defined by (3) will be denoted $m \colon \mathbb{R}^I \to \mathbb{R}^I$. Parameters $\theta$ are *uniquely* determined by $\mu$ by solving the equation system $\mu = m(\theta)$. Mean parameters are related with the log-partition function by

$$\frac{\mathrm{d}F(\theta)}{\mathrm{d}\theta} = m(\theta) \qquad (4)$$

Let $\phi(X) = \{\phi(x) \mid x \in X\}$ denote the range of $\phi$, a finite set of vectors from $\mathbb{R}^I$. The set of all realizable mean value vectors of $\phi$ is the convex hull of $\phi(X)$,

$$\operatorname{conv} \phi(X) = \Big\{ \sum_{x \in X} p(x)\phi(x) \,\Big|\, p(x) \ge 0,\ \sum_{x \in X} p(x) = 1 \Big\}$$

where $p$ stands for all possible distributions over $X$, not necessarily from the family. Every element of $\operatorname{conv} \phi(X)$ with $p(x) > 0$ (i.e., a strictly positive convex combination of $\phi$) can be obtained also as the mean of $\phi$ over a distribution *from* the family – thus, the range $m(\mathbb{R}^I)$ of the map $m$ is the interior of $\operatorname{conv} \phi(X)$.

### 2.2 Entropy and convex conjugacy

The entropy of distribution (1) as a function of $\theta$ equals $F(\theta) - \theta m(\theta)$. Let $H(\mu)$ denote the entropy of the distribution as a function of $\mu$. It is defined implicitly: we first take any $\theta$ satisfying $\mu = m(\theta)$ and then let $H(\mu) = F(\theta) - \theta\mu$. The function $H$ is positive and concave and its domain is the interior of $\operatorname{conv} \phi(X)$.

The functions $F$ and $-H$ are related by convex conjugacy (Legendre-Fenchel transform), which says that any $\mu$ from the interior of $\operatorname{conv} \phi(X)$ and any $\theta$ satisfy Fenchel's inequality

$$F(\theta) - H(\mu) - \theta\mu \ge 0 \qquad (5)$$

where equality holds if and only if $\mu = m(\theta)$. An alternative view of (5) is that $F(\theta) - H(\mu) - \theta\mu$ is the KL-divergence between the distribution determined by $\theta$ and the distribution determined by $\mu$.

Notice that the equality (4) can be obtained by minimizing the left-hand side of (5) with respect to $\theta$. Similarly, minimizing with respect to $\mu$ yields

$$\frac{\mathrm{d}H(\mu)}{\mathrm{d}\mu} = -\theta \qquad (6)$$

where $\theta$ is the (unique) solution of $m(\theta) = \mu$.

### 2.3 Reparameterizations

Now, suppose the basis functions $\phi_i$ are affinely dependent, that is, they form an overcomplete representation of the family. These dependencies can be written as

$$A\phi(x) = 0, \quad B\phi(x) = 1 \qquad \forall x \in X \qquad (7)$$

for some matrices $A$ and $B$, where 0 and 1 denote here column vectors of zeros and ones. Thus, matrix $A$ captures homogeneous dependencies and matrix $B$ captures inhomogeneous dependencies. It follows that

$$\mathrm{aff}\,\phi(X) = \{\,\mu \in \mathbb{R}^I \mid A\mu = 0,\ B\mu = 1\,\}$$

is the affine hull of the set $\phi(X)$.

Let $\alpha$ and $\beta$ be arbitrary row vectors and let

$$\theta' = \theta + \alpha A + \beta B \qquad (8)$$

Then, $\theta'\phi(x) = \theta\phi(x) + \beta 1$ and $F(\theta') = F(\theta) + \beta 1$. It follows that transformation (8) preserves distribution (1) and it is thus a *reparameterization* of the distribution. We will refer to the subclass of reparameterizations with $\beta = 0$ as *homogeneous reparameterizations*.

For an overcomplete representation, $\theta$ is no longer determined by $\mu$ uniquely but only up to reparameterizations, $m(\theta + \alpha A + \beta B) = m(\theta)$.

Moreover, equality (6) can no longer be used because the partial derivatives of $H(\mu)$ are undefined – only directional derivatives parallel to the space $\mathrm{aff}\,\phi(X)$ are defined. Let

$$\nabla_\nu H(\mu) = \lim_{t \to 0} \frac{H(\mu + t\nu) - H(\mu)}{t} = \left.\frac{\mathrm{d}H(\mu + t\nu)}{\mathrm{d}t}\right|_{t=0}$$

denote the directional derivative of $H(\mu)$ in direction $\nu \in \mathbb{R}^I$. To be parallel to $\mathrm{aff}\,\phi(X)$, $\nu$ has to satisfy $A\nu = 0$ and $B\nu = 0$. Then (6) generalizes to

$$\nabla_\nu H(\mu) = -\theta\nu \qquad \forall \nu\colon A\nu = 0,\ B\nu = 0 \qquad (9)$$

This is consistent with the fact that $\theta\nu$ is invariant to reparameterizations, $(\theta + \alpha A + \beta B)\nu = \theta\nu$.

## 3 Gibbs distribution

In §3, we show how the Gibbs distribution on a graphical model arises as a special exponential family.

Let $V$ be a set of variables. Let $E \subseteq 2^V$ be a set variable subsets, i.e., $(V, E)$ is a hypergraph[2]. We assume $E$ contains no one-element subsets. A variable $v$ takes states $x_v \in X_v$, where $X_v$ is a finite domain of the variable. For a hyperedge $a \in E$, let $X_a = \times_{v \in a} X_v$ denote the Cartesian product of domains of variables $a$. Elements of $X_a$ will be denoted $x_a$.

We instantiate $(X, I, \phi)$ such that distribution (1) becomes the Gibbs distribution on hypergraph $(V, E)$. Let $X = X_V$ be the Cartesian product of all variable domains. Let

$$I = \{(v, x_v) \mid v \in V,\ x_v \in X_v\} \cup \{(a, x_a) \mid a \in E,\ x_a \in X_a\}$$

[2]Though we consider the factor-graph BP in the paper, we do not use the concept of a factor graph – we use a hypergraph instead, which is clearly equivalent.

For $i \in I$, we denote the $i$-component of vector $\theta$ and $\mu$ by $\theta_v(x_v), \theta_a(x_a)$ and $\mu_v(x_v), \mu_a(x_a)$, respectively. Let $\phi\colon X \to \{0, 1\}^I$ be indicator functions chosen such that

$$\theta\phi(x) = \sum_{v \in V} \theta_v(x_v) + \sum_{a \in E} \theta_a(x_a) \qquad (10)$$

Now, distribution (1) is the Gibbs distribution and $\mu = m(\theta)$ are its marginals,

$$\mu_v(x_v) = \sum_{x_{V \setminus v}} p(x \mid \theta), \qquad \mu_a(x_a) = \sum_{x_{V \setminus a}} p(x \mid \theta)$$

The polytope $\mathrm{conv}\,\phi(X)$ contains all realizable marginal vectors $\mu$ and is known as the *marginal polytope* (Wainwright & Jordan, 2008). Moreover, for this choice of $(X, I, \phi)$ we have $\{0, 1\}^I \cap \mathrm{aff}\,\phi(X) = \phi(X)$.

### 3.1 Affine dependencies

Now we specify matrices $A$ and $B$, which capture affine dependencies among functions $\phi_i$. We do this indirectly, by writing down products $A\mu$, $B\mu$, $\alpha A$ and $\beta B$.

Equation systems $A\mu = 0$ and $B\mu = 1$ turn out to be the familiar marginalization and normalization conditions, respectively:

$$\sum_{x_{a \setminus v}} \mu_a(x_a) - \mu_v(x_v) = 0 \qquad (11\mathrm{a})$$

$$\sum_{x_v} \mu_v(x_v) = 1, \quad \sum_{x_a} \mu_a(x_a) = 1 \qquad (11\mathrm{b})$$

Let us remark that (11a) describes only a subset of all existing homogeneous dependencies among $\phi_i$, namely those that couple hyperedges with variables, and omits those that couple pairs of hyperedges. But this is a limitation of the factor-graph BP compared to the generalized BP. All existing dependences would be described by $\sum_{x_{a \setminus b}} \mu_a(x_a) = \sum_{x_{b \setminus a}} \mu_b(x_b)$. If $E \subseteq \binom{V}{2}$ is an ordinary graph (that means, there are only pairwise interactions), (11a) describes all existing dependencies.

Reparameterization $\theta' = \theta + \alpha A + \beta B$ reads

$$\theta'_v(x_v) = \theta_v(x_v) - \sum_{a \ni v} \alpha_{av}(x_v) + \beta_v \qquad (12\mathrm{a})$$

$$\theta'_a(x_a) = \theta_a(x_a) + \sum_{v \in a} \alpha_{av}(x_v) + \beta_a \qquad (12\mathrm{b})$$

Let us explain the detailed meaning of (12).

We define the *elementary homogeneous reparameterization* as follows: pick any pair $(a, v)$ with $v \in a$, subtract an arbitrary unary function $\alpha_{av}(\cdot)$ from function $\theta_v(\cdot)$, and add the same function to $\theta_a(\cdot)$:

$$\theta'_v(x_v) \leftarrow \theta_v(x_v) - \alpha_{av}(x_v) \qquad (13\mathrm{a})$$
$$\theta'_a(x_a) \leftarrow \theta_a(x_a) + \alpha_{av}(x_v) \qquad (13\mathrm{b})$$

Since $\alpha_{av}(x_v)$ cancels out, this preserves the sum $\theta_v(x_v) + \theta_a(x_a)$ and hence also the function (10). Applying transformations (13) to all pairs $(a, v)$ yields the terms with $\alpha$ in (12), i.e., the homogeneous reparameterization $\theta' = \theta + \alpha A$.

Reparameterization $\theta' = \theta + \beta B$ simply adds constants $\beta_v$, $\beta_a$ to all functions $\theta_v(\cdot)$, $\theta_a(\cdot)$.

Let us point out that papers on graphical models usually mean by 'reparameterizations' only homogeneous reparameterizations, or are not explicit about that.

Reparameterizations in the form (12) and (13) were first used by (Shlezinger, 1976) in LP relaxation of the problem $\max_{x \in X} \theta \phi(x)$ (i.e., finding modes of a Gibbs distribution). More can be found in modern revisions (Werner, 2007; Werner, 2010) of this approach.

## 4 Belief propagation

In the most general formulation (Yedidia et al., 2005), BP and related algorithms and free energies start with decomposing the original hypergraph into a collection of sub-hypergraphs (typically, hypertrees). Each sub-hypergraph is assigned a *counting number* (negative, zero, or positive) such that every hyperedge of the original hypergraph is counted exactly once in total.

In the factor-graph BP, our hypergraph $(V, E)$ is decomposed into the collection of sub-hypergraphs $E^v$ and $E^a$, where $v \in V$ and $a \in E$. Hypergraph $E^v$ contains only variable $v$. Hypergraph $E^a$ contains hyperedge $a$ and variables $v \in a$. The counting number of $E^a$ equals 1 and the counting number of $E^v$ equals $1 - n_v$, where $n_v = \sum_{a \ni v} 1$.

Each sub-hypergraph defines its own local Gibbs distribution. Let the distribution on $E^v$ and $E^a$ be denoted respectively by

$$p^v(x_v | \theta) = \exp\left[\theta_v(x_v) - F^v(\theta)\right] \quad (14a)$$
$$p^a(x_a | \theta) = \exp\left[\theta_a(x_a) + \sum_{v \in a} \theta_v(x_v) - F^a(\theta)\right] \quad (14b)$$

where the local log-partition functions read

$$F^v(\theta) = \bigoplus_{x_v} \theta_v(x_v) \quad (15a)$$
$$F^a(\theta) = \bigoplus_{x_a}\left[\theta_a(x_a) + \sum_{v \in a} \theta_v(x_v)\right] \quad (15b)$$

Similarly, the entropies of distributions (14) read[3]

$$H^v(\mu) = -\sum_{x_v} \mu_v(x_v) \log \mu_v(x_v) \quad (16a)$$
$$H^a(\mu) = -\sum_{x_a} \mu_a(x_a) \log \mu_a(x_a) \quad (16b)$$

---
[3]It might seem surprising that numbers $\mu_v(x_v)$ for $v \in a$ are absent in (16b). But (16b) is correct, variables really have zero counting numbers in hypergraph $E^a$.

Let us define two functions

$$\tilde{F}(\theta) = \sum_{v \in V}(1 - n_v) F^v(\theta) + \sum_{a \in E} F^a(\theta) \quad (17)$$
$$\tilde{H}(\mu) = \sum_{v \in V}(1 - n_v) H^v(\mu) + \sum_{a \in E} H^a(\mu) \quad (18)$$

While the function $\tilde{H}$ is the well-known Bethe entropy approximation, $\tilde{F}$ can be seen as the 'Bethe log-partition function'. To our knowledge, the function $\tilde{F}$ was not mentioned in previous works.

Next we proceed as follows. In §4.1 we define the BP algorithm and its fixed points. Then we give two interpretations of BP fixed points:

- In §4.2 we recall the well-known result by (Yedidia et al., 2000; Yedidia et al., 2005) that BP fixed points correspond to stationary points of the (negative) Bethe free energy $\theta\mu + \tilde{H}(\mu)$ on the space $\{\mu > 0 \mid A\mu = 0, \ B\mu = 1\}$. We refer to this as the *dual interpretation*.

- In §4.3 we present our main result, that BP fixed points correspond to stationary points of the function $\tilde{F}(\theta)$ on the space of homogeneous reparameterizations of $\theta$. We refer to this as the *primal interpretation*.

Here, we use the term 'stationary point' in a slightly broader meaning than is usual: a *stationary point of a function on an affine space* is a point where all directional derivatives parallel to that space vanish.

### 4.1 BP algorithm and its fixed points

Usually, BP is formulated in terms of passing messages, following (Pearl, 1988). We formulate it here in terms of reparameterizations. Our formulation is related to but different from (Wainwright et al., 2004).

In BP, probabilities (14) are seen as approximations of the true variable and hyperedge marginals of the Gibbs distribution (1). For a general $\theta$, they fail to satisfy the marginal consistency condition

$$\sum_{x_{a \setminus v}} p^a(x_a | \theta) = p^v(x_v | \theta) \quad (19)$$

which has to be satisfied by true marginals. The BP algorithm seeks to reparameterize $\theta$ such that (19) holds. Since functions (14) are invariant to reparameterizations $\theta' = \theta + \beta B$, only homogeneous reparameterizations can be considered. Plugging (14) into (19) yields

$$\bigoplus_{x_{a \setminus v}}\left[\theta_a(x_a) + \sum_{u \in a \setminus v} \theta_u(x_u)\right] = \text{const}_{av} \quad (20)$$

where $\text{const}_{av} = F^a(\theta) - F^v(\theta)$ are constants independent on $x_v$. We define a *BP fixed point* to be a vector $\theta$ satisfying (20).

A single update of the BP algorithm (its serial version) enforces condition (20) to hold for a single pair $(a,v)$ by applying the elementary homogeneous reparameterization (13) to the pair $(a,v)$. This determines $\alpha_{av}(\cdot)$ in (13) up to a constant. This constant is set so that $\bigoplus_{x_v} \alpha_{av}(x_v) = \bigoplus_{x_v} 0$, which ensures that numbers $\theta$ stay bounded during the algorithm.

In our exponential family formalism, the BP fixed point condition can be stated concisely as follows. Let a map $\mu = \tilde{m}(\theta)$ be defined by $\mu_v(x_v) = p^v(x_v|\theta)$, $\mu_a(x_a) = p^a(x_a|\theta)$. Map $\tilde{m}$ can be seen as an approximation of the true marginal map $m$. Now, BP fixed point condition (19) reads simply $A\tilde{m}(\theta) = 0$.

The true map $m$ satisfies $Am(\theta) = 0$, $Bm(\theta) = 1$ and $m(\theta + \alpha A + \beta B) = m(\theta)$. In contrast, $\tilde{m}$ satisfies only $B\tilde{m}(\theta) = 1$ and $\tilde{m}(\theta + \beta B) = \tilde{m}(\theta)$ in general. BP seeks to reparameterize $\theta$ such that also $A\tilde{m}(\theta) = 0$, i.e., to solve the system $A\tilde{m}(\theta + \alpha A) = 0$ for $\alpha$.

### 4.2 Dual interpretation of BP

In variational inference (Wainwright & Jordan, 2008), the log-partition function $F$ and marginals $m$ are computed indirectly via convex conjugacy between $F$ and $-H$. Fenchel's inequality (5) implies that

$$F(\theta) = \max\{\theta\mu + H(\mu) \mid \mu > 0,\ \mu \in \operatorname{conv}\phi(X)\} \quad (21)$$

where the optimum is attained at $\mu = m(\theta)$. This so far provides no advantage because both the marginal polytope $\operatorname{conv}\phi(X)$ and the entropy function $H$ are defined in an intractable way. The trick is to replace them with their tractable approximations. Then, the optimal argument and value of (21) is an approximation of the true $m(\theta)$ and $F(\theta)$, respectively.

If the polytope $\operatorname{conv}\phi(X)$ is approximated with the 'local polytope' (Wainwright & Jordan, 2008)

$$[0,1]^I \cap \operatorname{aff}\phi(X) = \{\mu \geq 0 \mid A\mu = 0,\ B\mu = 1\} \quad (22)$$

and the true entropy $H$ with the Bethe entropy (18), we obtain the Bethe variational problem

$$\max\{\theta\mu + \tilde{H}(\mu) \mid \mu > 0,\ A\mu = 0,\ B\mu = 1\} \quad (23)$$

where $-\theta\mu - \tilde{H}(\mu)$ is known as the Bethe free energy. In general, $[0,1]^I \cap \operatorname{aff}\phi(X) \supset \operatorname{conv}\phi(X)$ and $\tilde{H} \neq H$. However, if the factor graph of our graphical model is acyclic then $[0,1]^I \cap \operatorname{aff}\phi(X) = \operatorname{conv}\phi(X)$ and $\tilde{H} = H$. (Wainwright & Jordan, 2008; Yedidia et al., 2005).

Let us emphasize that the BP algorithm does not directly solve problem (23). BP maintains $\mu = \tilde{m}(\theta)$, which ensures $\mu > 0$ and $B\mu = 1$, and tries to reparameterize $\theta$ so that $A\mu = 0$. Thus, $\mu$ is infeasible to (23) until BP converges. Operating on the Lagrange multipliers of (23), BP is a *dual* algorithm to solve (23).

(Yedidia et al., 2005) showed that BP fixed points correspond to stationary points of problem (23). We need to say precisely what is meant by this correspondence because we defined BP fixed points in terms of $\theta$ and stationary points of (23) in terms of $\mu$. The correspondence is given by the map $\mu = \tilde{m}(\theta)$. This map is one-to-one up to adding constants to functions $\theta_v(\cdot), \theta_a(\cdot)$, i.e., up to reparameterizations $\theta \leftarrow \theta + \beta B$.

Moreover, notice that the objective of (23) is invariant to homogeneous reparameterizations because $(\theta + \alpha A)\mu = \theta\mu$ for feasible $\mu$.

With this understanding, we can state Yedidia's result.

**Theorem 1.** *If $\theta$ and $\mu$ correspond through $\mu = \tilde{m}(\theta)$, the following statements are equivalent:*

- $A\tilde{m}(\theta) = 0$, *i.e., $\theta$ is a BP fixed point.*
- $\mu$ *is a stationary point of $\theta\mu + \tilde{H}(\mu)$ on the set $\{\mu > 0 \mid A\mu = 0,\ B\mu = 1\}$.*

Let us remark that, by the discussion in §2.3, the second statement says that the directional derivative of $\theta\mu + \tilde{H}(\mu)$ vanishes in all directions parallel to aff $\phi(X)$:

$$\nabla_\nu \tilde{H}(\mu) = -\theta\nu \qquad \forall \nu\colon A\nu = 0,\ B\nu = 0 \quad (24)$$

### 4.3 Primal interpretation of BP

Here we present our main result, which can be concisely stated as follows: the BP algorithm tries to find a vector $\alpha$ such that the gradient of $\tilde{F}(\theta + \alpha A)$ with respect to $\alpha$ vanishes.

This gradient can be conveniently evaluated at $\alpha = 0$ without loss of generality since the gradient at $\alpha \neq 0$ can be recovered by replacing $\theta$ with $\theta + \alpha A$. Thus, we claim that $\theta$ is a BP fixed point if and only if

$$\left.\frac{\mathrm{d}\tilde{F}(\theta + \alpha A)}{\mathrm{d}\alpha}\right|_{\alpha=0} = A\frac{\mathrm{d}\tilde{F}(\theta)}{\mathrm{d}\theta} = 0 \quad (25)$$

where the first equality follows from the chain rule.

An alternative interpretation of condition (25) is that $\theta$ is a stationary point of function $\tilde{F}(\theta)$ on the space of homogeneous reparameterizations of $\theta$. Recall that this is the space of vectors $\theta + \alpha A$ for all possible $\alpha$. Condition (25) says that all directional derivatives parallel to this space vanish. Now we formulate our result.

**Theorem 2.** *The following statements are equivalent:*

- $A\tilde{m}(\theta) = 0$, *i.e., $\theta$ is a BP fixed point.*
- $A\,[\mathrm{d}\tilde{F}(\theta)/\mathrm{d}\theta] = 0$, *i.e., $\theta$ is a stationary point of $\tilde{F}(\theta)$ on the space of homogeneous reparameterizations of $\theta$.*

*Proof.* In the first part of the proof, we express the derivative (25) in terms of $\tilde{m}(\theta)$.

We begin by expressing the derivative $\mathrm{d}\tilde{F}(\theta)/\mathrm{d}\theta$ in terms of $\tilde{m}(\theta)$. Differentiating (17) yields

$$\frac{\partial \tilde{F}(\theta)}{\partial \theta_v(x_v)} = (1 - n_v)\frac{\partial F^v(\theta)}{\partial \theta_v(x_v)} + \sum_{a \ni v} \frac{\partial F^a(\theta)}{\partial \theta_v(x_v)} \quad (26\mathrm{a})$$

$$\frac{\partial \tilde{F}(\theta)}{\partial \theta_a(x_a)} = \frac{\partial F^a(\theta)}{\partial \theta_a(x_a)} \quad (26\mathrm{b})$$

Let us denote $\mu = \tilde{m}(\theta)$ for brevity. By (4), we have

$$\frac{\partial F^v(\theta)}{\partial \theta_v(x_v)} = \mu_v(x_v)$$

$$\frac{\partial F^a(\theta)}{\partial \theta_a(x_a)} = \mu_a(x_a) \qquad \frac{\partial F^a(\theta)}{\partial \theta_v(x_v)} = \sum_{x_{a\setminus v}} \mu_a(x_a)$$

Plugging this into (26) and some manipulations yields

$$\frac{\partial \tilde{F}(\theta)}{\partial \theta_v(x_v)} = \mu_v(x_v) + \sum_{a \ni v} \gamma_{av}(x_v) \quad (27\mathrm{a})$$

$$\frac{\partial \tilde{F}(\theta)}{\partial \theta_a(x_a)} = \mu_a(x_a) \quad (27\mathrm{b})$$

where we denoted $\gamma = A\mu$, i.e.,

$$\gamma_{av}(x_v) = \sum_{x_{a\setminus v}} \mu_a(x_a) - \mu_v(x_v)$$

By (11a), the components of (25) read

$$\left.\frac{\partial \tilde{F}(\theta + \alpha A)}{\partial \alpha_{av}(x_v)}\right|_{\alpha=0} = \sum_{x_{a\setminus v}} \frac{\partial \tilde{F}(\theta)}{\partial \theta_a(x_a)} - \frac{\partial \tilde{F}(\theta)}{\partial \theta_v(x_v)} \quad (28)$$

Plugging (27) into (28) finally yields

$$\left.\frac{\partial \tilde{F}(\theta + \alpha A)}{\partial \alpha_{av}(x_v)}\right|_{\alpha=0} = -\sum_{b \ni v,\, b \neq a} \gamma_{bv}(x_v) \quad (29)$$

In the second part of the proof, we express the two statements in Theorem 2 in terms of $\gamma$. The first statement is equivalent to system (30a) below. By (29), the second statement is equivalent to system (30b).

$$\gamma_{av}(x_v) = 0 \qquad \forall a \in E,\ v \in a,\ x_v \quad (30\mathrm{a})$$

$$\sum_{b \ni v,\, b \neq a} \gamma_{bv}(x_v) = 0 \qquad \forall a \in E,\ v \in a,\ x_v \quad (30\mathrm{b})$$

We need to show that systems (30a) and (30b) are equivalent. This can be shown separately for each pair $(v, x_v)$. Pick $(v, x_v)$ and write $\gamma_a$ instead of $\gamma_{av}(x_v)$ for simplicity. Then we need to show that an arbitrary set of numbers $\{\gamma_a \mid a \ni v\}$ satisfies the equivalence

$$\bigl[\gamma_a = 0\ \forall a \ni v\bigr] \iff \Bigl[\sum_{b \ni v,\, b \neq a} \gamma_b = 0\ \forall a \ni v\Bigr]$$

which is already easy. ∎

Next, we give a second order property of function $\tilde{F}$.

**Theorem 3.** *Consider $\tilde{F}(\theta + \alpha A)$ as a function of $\alpha$. Every stationary point of this function is a saddle point.*

*Proof.* We need to show that the Hessian

$$\frac{\mathrm{d}^2 \tilde{F}(\theta + \alpha A)}{\mathrm{d}\alpha^2}$$

is indefinite at any point $\alpha$ satisfying $A\tilde{m}(\theta + \alpha A) = 0$. It suffices to show that only a partial Hessian is indefinite. We obtain this partial Hessian by computing the partial derivatives $\partial^2 \tilde{F}(\theta + \alpha A)/\partial \alpha_k\, \partial \alpha_\ell$ only for some of all possible pairs $(k, \ell)$. After some work (we do not present details of the derivation) we get

$$\frac{\partial^2 \tilde{F}(\theta + \alpha A)}{\partial \alpha_{av}(x_v)\, \partial \alpha_{bv}(x_v)} = \begin{cases} 0 & \text{if } a = b \\ [\mu_v(x_v) - 1]\mu_v(x_v) & \text{if } a \neq b \end{cases}$$

where $\mu = \tilde{m}(\theta + \alpha A)$. This holds only if $A\mu = 0$, at points $A\mu \neq 0$ the derivative is more complex. The derivative takes only two values, depending on whether $a = b$ or $a \neq b$. Hence the diagonal elements of the partial Hessian are zero and the remaining elements are equal. Any such matrix is indefinite. ∎

### 4.4 Relation of primal and dual view

One can notice that $\tilde{F}$, $\tilde{H}$, $\tilde{m}$ are related by certain equalities, which can be seen as 'rudiments' of convex conjugacy relationship among the true $F$, $H$, $m$.

Thus, (27) shows that if $\theta$ is a BP fixed point then

$$\frac{\mathrm{d}\tilde{F}(\theta)}{\mathrm{d}\theta} = \tilde{m}(\theta) \quad (31)$$

In contrast to (4), equality (31) holds only at BP fixed points because of the extra term $\sum_{a \ni v} \gamma_{av}(x_v)$ in (27a), which vanishes at and only at BP fixed points. In fact, this might suggest that the map $\mu = \mathrm{d}\tilde{F}(\theta)/\mathrm{d}\theta$ is a more fundamental object than the map $\mu = \tilde{m}(\theta)$ – but we do not further pursue this observation here.

If $\mu = \tilde{m}(\theta)$ and $A\mu = 0$ (i.e., $\theta$ is a BP fixed point) then

$$\tilde{F}(\theta) - \tilde{H}(\mu) - \theta\mu = 0 \quad (32)$$

Unlike Fenchel's equality for true $F$, $H$, $m$, equality (32) fails to hold if $A\mu \neq 0$. Interestingly, we observed that condition $A\mu = 0$ becomes unnecessary if the form (18) of the Bethe entropy is replaced by its different form. Let

$$\tilde{H}(\mu) = \sum_{v \in V} H^v(\mu) - \sum_{a \in E} J^a(\mu) \quad (33)$$

where

$$J^a(\mu) = \sum_{x_a} \mu_a(x_a) \log \frac{\mu_a(x_a)}{\prod_{v \in a} \mu_v(x_v)}$$

is the KL-divergence between $\mu_a(x_a)$ and $\prod_{v \in a} \mu_v(x_v)$. Functions (18) and (33) are equal for $A\mu = 0$ but different otherwise (Wainwright & Jordan, 2008, §4.1.2). It can be easily verified that with this form of $\tilde{H}$, equality (32) holds for $\mu = \tilde{m}(\theta)$ even if $A\mu \neq 0$. In other words, substitution $\mu = \tilde{m}(\theta)$ transforms the function $\theta\mu + \tilde{H}(\mu)$ into $\tilde{F}(\theta)$.

The Bethe entropy has a clear meaning: for acyclic graphs, $\tilde{H}$ equals the true entropy $H$. It follows from (32) that $\tilde{F}$ has a similar property: for acyclic graphs, $\tilde{F}$ equals the true log-partition function $F$ but only if $A\tilde{m}(\theta) = 0$ (i.e., only on the space of BP fixed points).

## 5 Conclusion

We have presented a novel interpretation of loopy belief propagation. While it was known that BP fixed points correspond to stationary points of the Bethe free energy on the local polytope, we have shown that they also correspond to stationary points of the 'Bethe log-partition function' on the space of homogeneous reparameterizations. To the best of our knowledge, this simple observation was not made before. The two interpretations are exactly complementary – however, they are not related by classical convex duality because function $\tilde{H}$ is not concave and $\tilde{F}$ is not convex.

So far, BP was understood as an algorithm to seek for a common zero of a set of explicitly unrelated equations. Our result shows that these equations are partial derivatives of the single function $\tilde{F}(\theta + \alpha A)$ of $\alpha$ without any additional constraints.

One would expect that finding a stationary point of a single multivariate analytic function must be easier than solving a system of unrelated non-linear equations – but this is true only if the stationary point is a local extreme. Unfortunately, all stationary points of the function $\tilde{F}(\theta + \alpha A)$ are saddle points, and finding a saddle point can be much harder (and little literature seems to exist about it). Therefore, we currently do not know whether our result can provide new insights into (non-)convergence of BP.

Various generalized versions of BP are often designed via dual considerations involving local free energies and entropies. Our result suggests that free energies may not be needed for this at all, the primal route via reparameterizations and local log-partition functions may be simpler. This is open to future research.

Although we have not demonstrated any practical consequences of our contribution, we believe that the presented mathematical framework, which treats reparameterizations explicitly and incorporates them into the exponential family language, brings more clarity in the theoretical understanding of graphical models.

**Acknowledgements**

This research was supported by the European Commission grant 215078 (DIPLECS) and the Czech government grant MSM6840770038.

## References

Heskes, T. (2003). Stable fixed points of loopy belief propagation are minima of the Bethe free energy. *Advances in Neural Information Processing Systems (NIPS)* (pp. 359–366).

Heskes, T. (2006). Convexity arguments for efficient minimization of the Bethe and Kikuchi free energies. *Jr. of Artificial Intelligence Research, 26*, 153–190.

Kschischang, F. R., Frey, B. J., & Loeliger, H. A. (2001). Factor graphs and the sum-product algorithm. *IEEE Trans. Inf. Theory, 47*, 498–519.

Pearl, J. (1988). *Probabilistic reasoning in intelligent systems: Networks of plausible inference*. San Francisco: Morgan Kaufmann.

Shlezinger, M. I. (1976). Syntactic analysis of two-dimensional visual signals in noisy conditions. *Cybernetics and Systems Analysis, 12*, 612–628. Translation from Russian.

Wainwright, M., Jaakkola, T., & Willsky, A. (2004). Tree consistency and bounds on the performance of the max-product algorithm and its generalizations. *Statistics and Computing, 14*, 143–166.

Wainwright, M. J., & Jordan, M. I. (2008). Graphical models, exponential families, and variational inference. *Foundations and Trends in Machine Learning, 1*, 1–305.

Werner, T. (2007). A linear programming approach to max-sum problem: A review. *IEEE Trans. Pattern Analysis and Machine Intelligence, 29*, 1165–1179.

Werner, T. (2010). Revisiting the linear programming relaxation approach to Gibbs energy minimization and weighted constraint satisfaction. *IEEE Trans. Pattern Analysis and Machine Intelligence*. To appear in August 2010.

Yedidia, J., Freeman, W. T., & Weiss, Y. (2000). Generalized belief propagation. *Neural Information Processing Systems (NIPS)* (pp. 689–695).

Yedidia, J. S., Freeman, W. T., & Weiss, Y. (2005). Constructing free-energy approximations and generalized belief propagation algorithms. *IEEE Trans. Information Theory, 51*, 2282–2312.